\PassOptionsToPackage{table}{xcolor} 
\documentclass[11pt]{article}

\usepackage[preprint]{acl}
\usepackage{mdframed}
\usepackage{tcolorbox}
\usepackage{times}
\usepackage{latexsym}
\usepackage{amsmath}
\usepackage{amssymb}
\usepackage{booktabs} 
\usepackage [table]{xcolor} 
\usepackage[T1]{fontenc}
\usepackage[utf8]{inputenc}
\usepackage{newunicodechar} 
\newunicodechar{’}{\textquoteright} 
\newunicodechar{—}{\textemdash} 

\usepackage{microtype}

\usepackage{inconsolata}

\usepackage{graphicx}

\title{Taking the \textsc{PUlSE} of Medical Reasoning: Step-Level Retrieval Timing for Knowledge-Grounded Inference}

\author{
  Danyang Zhao,
  Sendong Zhao\thanks{
    Corresponding author: \texttt{sdzhao@ir.hit.edu.cn}.
  },
  Yanrui Du,
  Haochun Wang,
  Ming Ma \\
  \textbf{
    Wenhao Li,
    Jingbo Wang,
    Jiatong Liu,
    Bing Qin
  } \\
  Harbin Institute of Technology
}

\begin{document}
\maketitle


\begin{abstract}
Reliable medical reasoning requires LLMs to dynamically access external biomedical evidence to improve factual trustworthiness and reasoning accuracy. 
However, indiscriminate retrieval can introduce noise and latency, raising a critical retrieval-timing problem: at each reasoning step, the system must decide whether the model's internal knowledge is reliable or external evidence is needed. 
We propose \textbf{\textsc{PUlSE}}, a lightweight probing method that uses the model’s internal representations as intrinsic signals for step-level retrieval control. 
Our posterior trajectory analysis reveals that erroneous steps exhibit substantially higher inter-layer volatility than correct ones, suggesting unstable internal knowledge. 
Based on this finding, \textsc{PUlSE} trains a compact probe on inter-layer geometric features to detect factual errors upon step completion, and uses it as an adaptive trigger for retrieval and step regeneration. 
Across four medical benchmarks, \textsc{PUlSE}-timed retrieval matches exhaustive per-step retrieval in accuracy while reducing retrieval calls by 72\%, and can transfer to an unseen legal domain, indicating potential domain-agnostic generality.
Overall, our study provides an initial step toward shifting retrieval-trigger design from surface-level confidence estimates to internal activation dynamics for adaptive access.
\end{abstract}

\section{Introduction}

Reliable medical reasoning requires Large Language Models (LLMs) to access external biomedical evidence to support factual trustworthiness and reasoning accuracy, particularly when specialized or rapidly evolving knowledge is needed~\citep{article,singhal2023large,bayat2023fleek,Berger2025ReasoningLI}.
Existing retrieval-augmented generation (RAG) research has devoted substantial attention to improving the relevance and quality of external evidence~\citep{medrag}.
However, in multi-step medical reasoning, factual errors may arise at specific steps rather than throughout the entire reasoning chain~\citep{wu2025medreason,sun2025reasonmed}, raising a question beyond evidence quality: how can retrieval timing be determined to achieve a better balance between reasoning accuracy and retrieval frequency than indiscriminate retrieval?

\begin{figure}[t]
    \centering
    \includegraphics[width=\columnwidth]{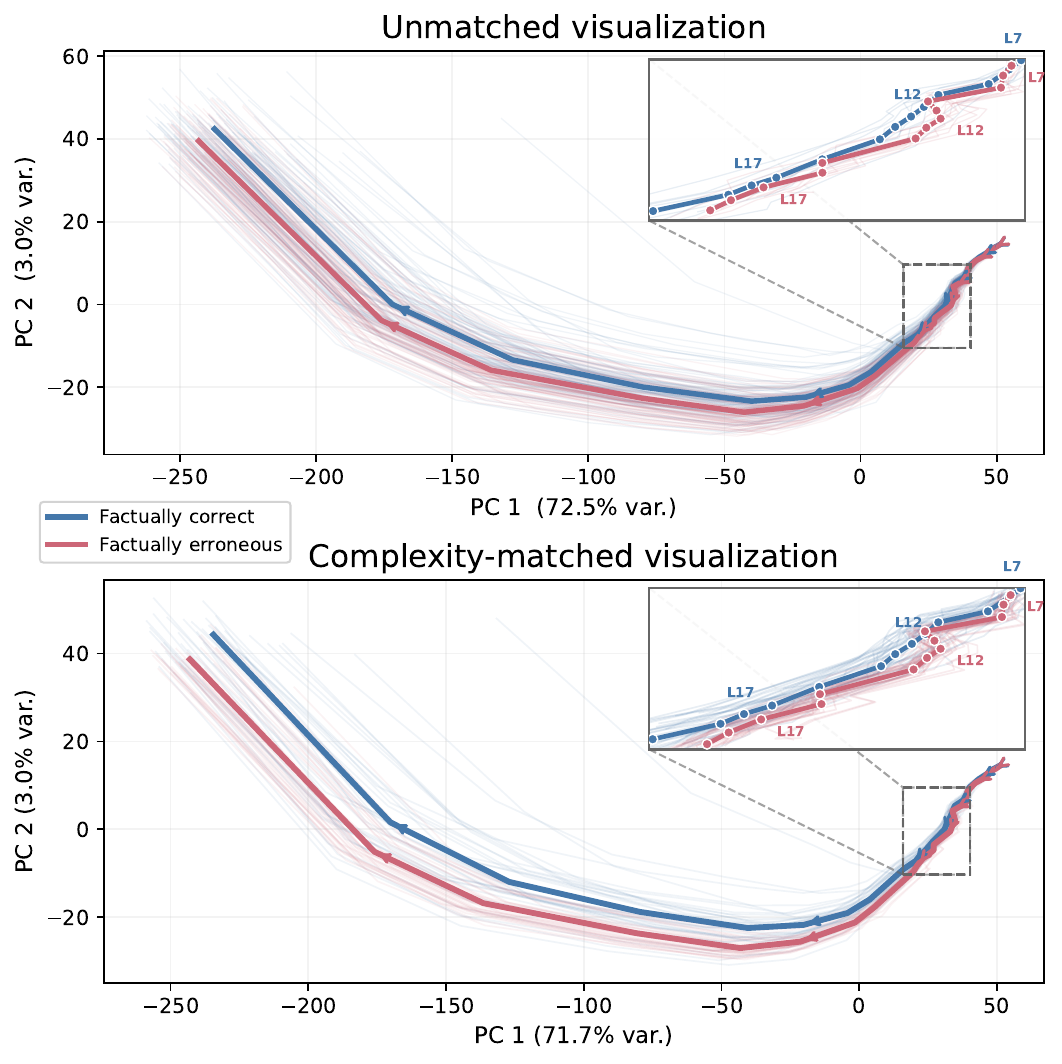}
    \caption{Layer-wise PCA trajectories of hidden states. Bold lines: group means; faint lines: individual samples. (a) Unmatched. (b) Complexity-matched.}
    \label{fig:motivation}
\end{figure}

Existing retrieval-augmented methods differ mainly in the granularity at which retrieval is triggered. 
Question-level retrieval retrieves evidence before reasoning begins~\citep{lewis2020retrieval,Lu2025DoctorRAGMR}, but is too coarse to address factual errors that emerge only in intermediate reasoning steps. 
Exhaustive step-level retrieval can provide targeted support, but requires frequent retrieval calls and may introduce unnecessary or distracting evidence~\citep{Huang2024ASO}. 
Token-level retrieval further refines the trigger granularity~\citep{jiang2023active,su-etal-2024-dragin}, yet it fails to preserve the semantic integrity of a complete reasoning step. 
Step-level adaptive retrieval therefore offers a natural granularity~\citep{yao-etal-2025-seakr}, and its effectiveness hinges on a critical retrieval-timing problem: after each reasoning step, the system must decide whether the model's internal knowledge is reliable or external evidence is required.

A central challenge is how to obtain reliable trigger signals. 
Most existing approaches rely on output-level uncertainty signals, such as perplexity~\citep{si2022prompting}, entropy~\citep{kuhn2023semantic}, or maximum softmax probability, to estimate whether the model is likely to be wrong. 
While useful, these signals are derived from the final output distribution and may provide only a surface-level view of the model's reasoning state. 
Another line of work has shown that factual knowledge and reasoning-related information are encoded in the model's hidden representations~\citep{Meng2022LocatingAE,geva-etal-2021-transformer,azaria-mitchell-2023-internal,11456255}, and that probes over multi-layer representations can help assess reasoning quality~\citep{Ni2025ReProbeET}. 
The key is to translate these representations into a signal for deciding when to retrieve.
These findings suggest that layer-wise internal dynamics may contain richer trigger signals.

Motivated by this perspective, we investigate whether the layer-wise evolution of hidden states can serve as an intrinsic signal for step-level factual error detection. Our posterior trajectory analysis reveals that factually erroneous reasoning steps exhibit substantially higher inter-layer volatility than correct ones, even after controlling for reasoning complexity, as shown in Figure~\ref{fig:motivation}. This pattern suggests that when a model fails to ground a step in its internal knowledge, its representations undergo unstable transitions across layers. We term this phenomenon \emph{internal knowledge instability} and build upon it to propose \textsc{PUlSE} (\textbf{P}robing \textbf{U}ngrounded reasoning via \textbf{L}ayer-wise \textbf{S}tate \textbf{E}volution), a lightweight probing method for adaptive retrieval control. \textsc{PUlSE} extracts compact geometric features from hidden-state trajectories and detects factual errors upon step completion. Rather than retrieve at every step, it selectively triggers biomedical retrieval and step regeneration when a potential factual error is detected, reducing unnecessary retrieval calls. 

Our study evaluates \textsc{PUlSE} on four medical QA benchmarks. 
\textsc{PUlSE}-timed retrieval matches exhaustive per-step retrieval in downstream accuracy while reducing retrieval calls by 72\%, achieving a favorable balance between accuracy and retrieval frequency. 
Moreover, the same trigger transfers zero-shot to an unseen legal domain, suggesting that layer-wise trajectory volatility may capture a domain-agnostic signal of factual grounding failure rather than dataset-specific artifacts. 
Overall, our contributions can be summarized as follows:
\begin{itemize}
    \item \textbf{Geometric Signal.} We identify elevated inter-layer volatility in hidden states as a geometric signature of step-level factual errors, providing an intrinsic basis for retrieval timing.
    \item \textbf{Retrieval Trigger.} We propose \textsc{PUlSE}, a lightweight probe that determines retrieval timing from the geometric evolution of hidden states across layers. 
    \item \textbf{Retrieval Efficiency.} We show that \textsc{PUlSE}-timed retrieval matches exhaustive retrieval with 72\% fewer calls and generalizes zero-shot across domains. 
\end{itemize}

\section{Related Work}

Retrieval-Augmented Generation (RAG) mitigates LLM factual deficiencies by incorporating external knowledge~\citep{lewis2020retrieval}. A central challenge in multi-step reasoning is deciding \emph{when} to retrieve and \emph{how} to retrieve—exhaustive per-step retrieval introduces latency and noise, while omitting retrieval at knowledge-deficient steps allows errors to propagate.
\paragraph{Retrieval Strategy.}
To handle multi-step reasoning, methods such as IRCoT~\citep{trivedi-etal-2023-interleaving} and ReAct~\citep{ac1f09077393404a8bea5141d8710259} interleave retrieval with generation, while recent work improves query formulation and chunk selection~\citep{su-etal-2024-dragin}. These advances address the retrieval pipeline itself and are orthogonal to our contribution; \textsc{PUlSE} can serve as a front-end trigger for any such retriever.
\begin{figure*}
    \centering
    \includegraphics[width=\textwidth]{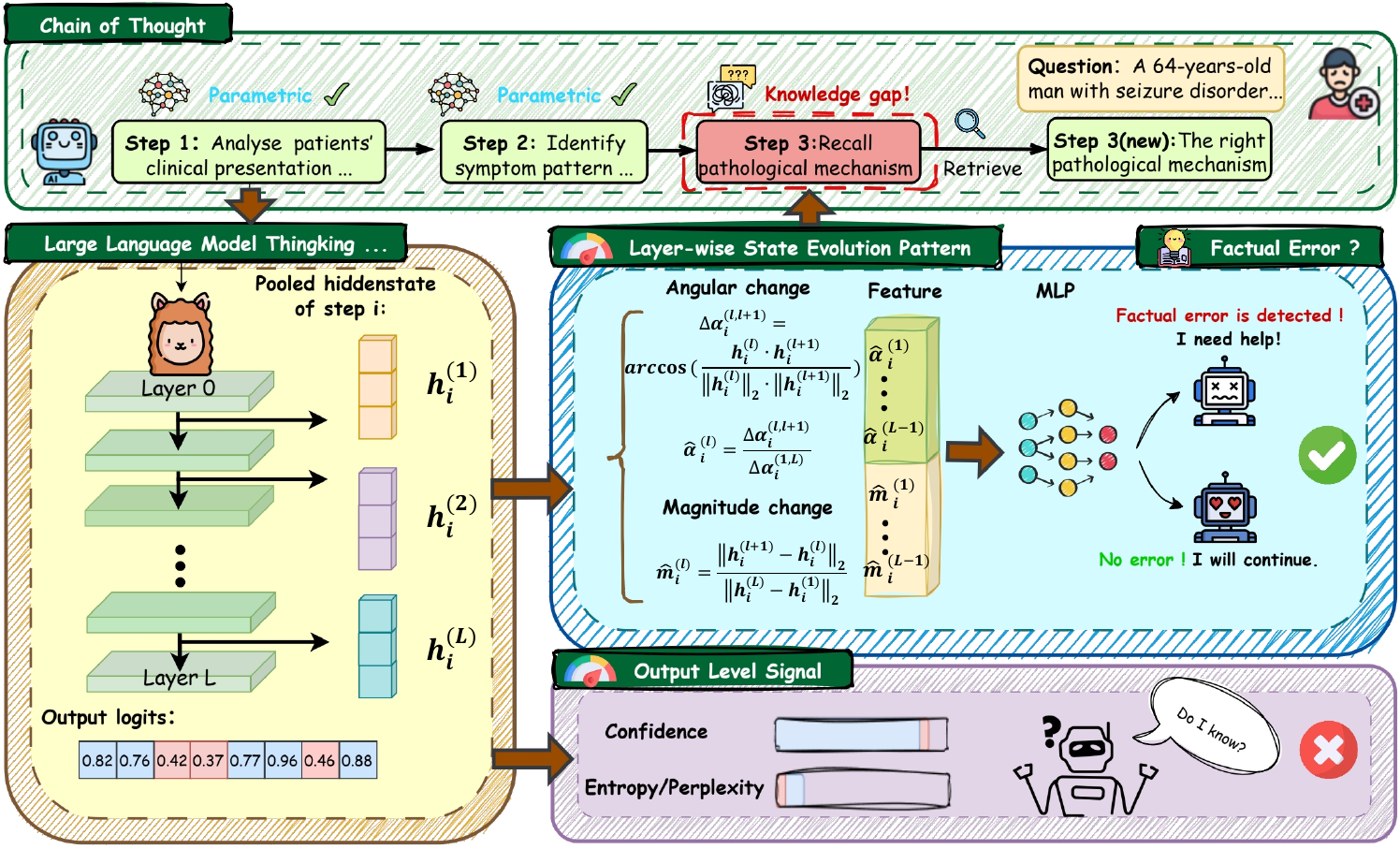}
    \caption{Overview of \textsc{PUlSE}. Layer-wise geometric trajectory features of each reasoning step are fed into an MLP probe to detect factual errors. Flagged steps trigger retrieval, whose accuracy gain interventionally validates the detection. Output-level signals (purple) serve as baselines.}
    \label{fig:method}
\end{figure*}
\paragraph{Retrieval Timing.}
The more fundamental question is \emph{when} retrieval is needed~\cite{Tang2025Eigen1AM}. Existing triggering signals fall into two categories by their source.

At the output level, methods rely on generation-time statistics: token probability thresholds~\citep{jiang2023active}, semantic entropy over sampled outputs~\citep{kuhn2023semantic}, logit-based uncertainty regressors~\citep{liu2024uncertainty}, and composite heuristics combining token probability with attention-based importance scoring~\citep{su-etal-2024-dragin}. Trained reflection tokens~\citep{asai2024self} also operate in the output space. These signals are cheap to compute but remain surface-level proxies that conflate linguistic uncertainty with genuine knowledge deficiency.

At the internal level, recent work examines hidden-state geometry for factual assessment. SeaKR~\citep{yao-etal-2025-seakr} uses hidden-state uncertainty to trigger retrieval adaptively. CoE~\citep{wang2025latent} aggregates cross-layer geometric feature changes into scalar scores, while \citet{cheang2025large} and \citet{lee2025the} analyze internal state geometry during knowledge recall and self-verification. However, these efforts operate at the granularity of a complete query or response—none localizes knowledge deficiency to individual reasoning steps.

\textsc{PUlSE} bridges this gap by bringing layer-wise geometric trajectory analysis to the step level. It learns how trajectories co-evolve across layers, capturing discriminative evolution patterns that signal factual errors within each reasoning step and providing fine-grained, domain-transferable retrieval triggering signals.

\section{Checking the \textsc{PUlSE}}
\label{sec:detection}

We propose \textsc{PUlSE} (Probing Ungrounded reasoning via Layer-wise State Evolution), a method that detects factual errors at individual reasoning steps by exploiting the geometric structure of layer-wise hidden-state trajectories.
Figure~\ref{fig:method} illustrates the overall workflow.

\subsection{Posterior Trajectory Analysis}
\label{sec:post_analysis}

To understand how a model's internal representations differ between factually correct and erroneous reasoning steps, we examine the layer-wise evolution of hidden states during chain-of-thought generation.

\paragraph{Step-level trajectory construction.}
Consider a reasoning step~$s_i$ spanning tokens $(t_i^{(1)}, \dots, t_i^{(K_i)})$. For a Transformer with $L$ layers, let $\mathbf{z}_{i,k}^{(l)} \in \mathbb{R}^d$ denote the hidden state of token~$t_i^{(k)}$ at layer~$l$. We obtain a single vector summarizing the step at each layer via mean pooling:
\begin{equation}
  h_i^{(l)} = \frac{1}{K_i} \sum_{k=1}^{K_i} \mathbf{z}_{i,k}^{(l)}
\end{equation}
The ordered sequence $(h_i^{(1)}, \dots, h_i^{(L)})$ then constitutes the \emph{layer-wise trajectory} of step~$s_i$, encoding how the model's representation of that step transforms as it propagates through successive layers.

\paragraph{Trajectory visualization.}
Each step is annotated as factually correct or erroneous (annotation protocol detailed in Appendix~\ref{appendix:annotation_protocol}).
We project these layer-wise representations into two dimensions via PCA and trace each step's trajectory across layers (Figure~\ref{fig:motivation}, top). Two patterns emerge clearly: (1)~trajectories of factually erroneous steps exhibit markedly higher volatility---particularly in mid-to-late layers---compared to correct steps, and (2)~the two classes show separation in the projected space. Crucially, this separation is not attributable to any single layer's activation; rather, it arises from the \emph{sequential pattern} of how representations evolve across the layer stack.

\paragraph{Controlling for step complexity.}
A natural concern is that the observed volatility merely reflects differences in reasoning complexity rather than factual correctness. To control for this confound, we stratify steps by token count and internal line-break structure as coarse complexity proxies, then repeat the PCA visualization within matched strata (Figure~\ref{fig:motivation}, bottom). The mid-to-late-layer volatility gap between correct and erroneous steps persists across all strata, confirming that the geometric divergence is not reducible to step complexity.

\paragraph{Implication.}
These observations indicate that layer-wise evolution carries an intrinsic, complexity-independent signal for distinguishing grounded from ungrounded reasoning, motivating the detection method we formalize next.

\subsection{Prior Error Detection}
\label{sec:priori_detection}

The posterior analysis reveals that factual errors leave a distinctive geometric footprint in layer-wise trajectories. To quantify trajectory volatility in a form amenable to learned classification, we decompose each step's trajectory into two complementary feature sequences that capture how the representation evolves across layers, then train a lightweight probe on these features.
\paragraph{Angular change sequence.}
We measure directional reorientation between adjacent layers via the normalized inter-layer angle:
\begin{equation}
  \hat{\alpha}_i^{(l)}
    = \frac{\alpha_i^{(l,\, l+1)}}
           {\alpha_i^{(l_s,\, l_e)}},
  \quad l = l_s, \dots, l_e{-}1
\end{equation}
where
\begin{equation}
  \alpha_i^{(a,b)}
    = \arccos\!\left(
        \frac{h_i^{(a)} \cdot h_i^{(b)}}
             {\|h_i^{(a)}\|_2 \;\|h_i^{(b)}\|_2}
      \right)
\end{equation}
This yields $A_i = (\hat{\alpha}_i^{(l_s)}, \dots, \hat{\alpha}_i^{(l_e-1)}) \in \mathbb{R}^{l_e - l_s}$; larger values indicate more pronounced directional shifts at that layer transition. Normalizing by the endpoint angle $\alpha_i^{(l_s, l_e)}$ removes the influence of a step's overall angular span.

\paragraph{Magnitude change sequence.}
Complementarily, we quantify how the representation scale evolves across successive layers by measuring the normalized inter-layer displacement:
\begin{equation}
  \Delta m_i^{(l)}
    = \frac{\| h_i^{(l+1)} - h_i^{(l)} \|_2}
           {\| h_i^{(l_e)} - h_i^{(l_s)} \|_2},
  \quad l = l_s, \dots, l_e{-}1
\end{equation}
This yields the sequence $M_i = (\Delta m_i^{(l_s)}, \dots, \Delta m_i^{(l_e-1)}) \in \mathbb{R}^{l_e - l_s}$, where each element captures the relative expansion or contraction at a single layer transition. Normalizing by the total displacement $\| h_i^{(l_e)} - h_i^{(l_s)} \|_2$ ensures comparability across steps of different overall magnitude.

\paragraph{Feature concatenation and probe.}
Together, angular and magnitude sequences provide a structured geometric characterization of trajectory volatility---the former tracking ``in what direction'' it turns, the latter tracking ``how far'' the representation moves. We concatenate them into a unified feature vector:
\begin{equation}
  \mathbf{f}_i = [A_i;\; M_i] \in \mathbb{R}^{2(l_e - l_s)}
\end{equation}
Here $[l_s, l_e]$ denotes a contiguous layer range restricted to mid-to-late layers, motivated by Section~\ref{sec:post_analysis}; its specific boundaries are validated in the ablation study (Section~\ref{sec:design}). An MLP probe $g\colon \mathbb{R}^{2(l_e - l_s)} \to [0,1]$ with two hidden layers (dimensions 128 and 64, ReLU activations) maps the trajectory feature vector to an error probability, trained with binary cross-entropy loss for up to 1{,}000 iterations using Adam. At inference time, given a new step~$s_i$, we extract its hidden states across $[l_s, l_e]$, compute $\mathbf{f}_i$, and obtain $g(\mathbf{f}_i)$; steps exceeding a threshold are flagged as likely factual errors.

\subsection{Detection Experiments}
\label{sec:probe_exp}

\begin{table*}[t]
\centering
\caption{Factual error detection performance on Qwen2.5 and Llama3.1. Threshold-independent metrics (AUROC, PR AUC) are reported as primary metrics. Baseline thresholds are selected via Youden's J statistic; probe-based methods use a default threshold of 0.5. Best results per dataset are in \textbf{bold}.}
\label{tab:probing}
\resizebox{\textwidth}{!}{%
\begin{tabular}{l *{10}{c}}
\hline
&\multicolumn{5}{c}{\textbf{MedMCQA (ID)}} &\multicolumn{5}{c}{\textbf{MMLU-Pro (Biology) (OOD)}} \\
\cline{2-6} \cline{7-11}
\textbf{Method}
& Prec. & Rec. & F1 & AUROC & PR AUC
& Prec. & Rec. & F1 & AUROC & PR AUC \\
\hline
\multicolumn{11}{l}{\emph{Backbone: Qwen2.5-7B-Instruct}} \\
\hline
PPL          & .4433 & .6160 & .5155 & .6112 & .4768 & .4643 & .0138 & .0268 & .4776 & .2200 \\
MaxProb      & .4480 & .5436 & .4912 & .6043 & .4727 & .5417 & .0138 & .0269 & .4838 & .2224 \\
Entropy      & .4538 & .6289 & .5272 & .6264 & .4932 & .2727 & .0477 & .0812 & .4695 & .2177 \\
CoE-R        & .4419 & .5436 & .4875 & .5843 & .4204 & .2774 & .6140 & .3822 & .5981 & .2808 \\
CoE-C        & .4571 & .7607 & .5710 & .6560 & .4948 & .3032 & .6394 & .4113 & .6442 & .3630 \\
ReProbe      & .4420 & \textbf{.9325} & .5998 & .7350 & \textbf{.5847} & .3287 & .6300 & .4320 & .6789 & .3866 \\
\rowcolor{green!10}
\textsc{PUlSE}         & \textbf{.5027} & .8571 & \textbf{.6337} & \textbf{.7385} & .5666 & \textbf{.3608} & \textbf{.6829} & \textbf{.4721} & \textbf{.7273} & \textbf{.4316} \\
\hline
\multicolumn{11}{l}{\emph{Backbone: Llama3.1-8B-Instruct}} \\
\hline
PPL          & .3306 & .3015 & .3154 & .5310 & .3185 & \textbf{.6352} & .2741 & .3856 & .6090 & .5156 \\
MaxProb      & .3822 & .1854 & .2497 & .5299 & .3141 & \textbf{.6352} & .2793 & .3830 & .6093 & .5149 \\
Entropy      & .3719 & .2772 & .3176 & .5446 & .3428 & .5634 & .2993 & .3909 & .5921 & .4836 \\
CoE-R        & .2964 & \textbf{.9757} & .4546 & .5323 & .2704 & .5683 & .5471 & .5575 & .7204 & .6341 \\
CoE-C        & .3047 & .7697 & .4365 & .6018 & .3552 & .5274 & .6575 & .5853 & .7257 & .6144 \\
ReProbe      & \textbf{.4474} & .6211 & .5201 & .7431 & .5103 & .6223 & .5830 & .6021 & .7406 & .6451 \\
\rowcolor{green!10}
\textsc{PUlSE}         & .4450 & .6816 & \textbf{.5385} & \textbf{.7528} & \textbf{.5150} & .6060 & \textbf{.6844} & \textbf{.6428} & \textbf{.7933} & \textbf{.6982} \\
\hline
\end{tabular}
}
\end{table*}

\paragraph{Datasets.}
We train the probe on 1{,}000 MedMCQA~\cite{pal2022medmcqa} and 173 AfriMedQA~\cite{nimo-etal-2025-afrimed} questions (70/30 train/val split; labeling protocol and step-level statistics in Appendix~\ref{appendix:data}).
We evaluate on an in-distribution set of 300 held-out MedMCQA questions and an out-of-distribution set from MMLU-Pro (Biology)~\cite{wang2024mmlu}, testing whether the geometric signal transfers without re-training. The resulting label distribution is imbalanced, motivating the use of PR\,AUC alongside AUROC.

\paragraph{Baselines.}
We compare against training-free output-level indicators
(PPL, MaxProb, Entropy) and hidden-state-based methods
(CoE-R/C~\citep{wang2025latent}, ReProbe~\citep{Ni2025ReProbeET}).
Full descriptions are provided in Appendix~\ref{appendix:baselines}.

\paragraph{Implementation details.}
We conduct experiments on Qwen2.5-7B-Instruct~\cite{yang2024qwen2} and Llama3.1-8B-Instruct~\cite{grattafiori2024llama}. We extract hidden states from layers 7--25 (Qwen) and 6--31 (Llama), and compute the magnitude and angular change sequences in Section~\ref{sec:priori_detection}. Layer intervals are selected on the MedMCQA training-validation split (Section~\ref{sec:design}). We report AUROC, PR\,AUC, F1, Precision, and Recall. 

\paragraph{Results and analysis.}
Table~\ref{tab:probing} reports detection performance on both backbones.
On the in-distribution MedMCQA set, \textsc{PUlSE} achieves the highest
AUROC on both backbones (0.7385 on Qwen, 0.7528 on Llama) and
consistently leads in F1. Compared with ReProbe, which concatenates all
layer states without modeling their order, the gain confirms that
preserving layer-by-layer geometric structure provides additional
discriminative signal.

The advantage widens on the out-of-distribution MMLU-Biology set: on
Llama, AUROC reaches 0.7933 versus 0.7406 for ReProbe (+5.3\,pp)
without any re-training, indicating that the learned patterns reflect a
general internal signature of factual errors rather than dataset-specific artifacts. Furthermore, the probe remains predictive after excluding steps with explicit uncertainty expressions (Appendix~\ref{appendix:non_hedging}).
Output-level methods (PPL, MaxProb, Entropy) degrade substantially under distribution shift, suggesting that surface-level confidence cannot reliably reveal the model's epistemic state regarding factual grounding.
\section{Retrieving on \textsc{PUlSE}}
\label{sec:retrieval}
Having established that \textsc{PUlSE} reliably detects step-level factual errors from trajectory geometry, we now validate its practical utility through an adaptive retrieval framework: if the probe correctly pinpoints where factual errors occur, supplying external evidence at those steps should yield greater accuracy gains with fewer retrievals.
\subsection{Error-Triggered Retrieval Framework}
\label{sec:framework}
Using the probe output as a retrieval trigger, we construct a
step-level adaptive retrieval framework. At inference time, the
model is prompted to generate chain-of-thought reasoning in which
each step begins with a ``Step'' prefix, yielding well-delimited semantic units.
After each step~$s_i$ is generated, we extract its layer-wise
trajectory, compute the feature vector~$\mathbf{f}_i$, and pass
it to the probe~$g$. If $g(\mathbf{f}_i) > \tau$ , where $\tau$ is the detection threshold, the system extracts important tokens from the factually erroneous step and retrieves relevant passages from an external
corpus, and then injects them into the context for subsequent reasoning;
otherwise, the model continues with its parametric knowledge alone,
avoiding unnecessary external evidence.

We explore two knowledge injection strategies, both of which regenerate
the flagged step while keeping all prior steps intact:
\textbf{(i)~\textsc{PUlSE}\textsubscript{front}}: retrieved passages are
prepended before the question following IRCoT~\citep{trivedi-etal-2023-interleaving};
\textbf{(ii)~\textsc{PUlSE}\textsubscript{step}}: retrieved passages are
inserted immediately before the flagged step.
\subsection{Retrieval Experiments}
\label{sec:retrieval_expe}

\begin{table*}[t]
\centering
\caption{Main results on four medical QA benchmarks using BM25 retrieval.
Acc: accuracy (\%). \#Ret: average number of retrieval calls per test instance.
\textbf{Bold}: best; \underline{underline}: second best within each backbone group (excluding CoT for \#Ret).}
\label{tab:main}
\renewcommand{\arraystretch}{1.0}
\setlength{\tabcolsep}{3pt}
\begin{tabular}{l cc cc cc cc}
\hline
 & \multicolumn{2}{c}{\textbf{MedMCQA}}
 & \multicolumn{2}{c}{\textbf{JAMA}}
 & \multicolumn{2}{c}{\textbf{MMLU-Pro (Health)}}
 & \multicolumn{2}{c}{\textbf{MMLU-Pro (Biology)}} \\
\cline{2-3} \cline{4-5} \cline{6-7}\cline{8-9}
\textbf{Method}
  & Acc (\%) & \#Ret\,$\downarrow$
  & Acc (\%) & \#Ret\,$\downarrow$
  & Acc (\%) & \#Ret\,$\downarrow$
  & Acc (\%) & \#Ret\,$\downarrow$ \\
\hline
\multicolumn{9}{l}{\emph{Backbone: Qwen2.5-7B-Instruct}} \\
\hline
CoT
  & 57.00 & 0
  & 50.30 & 0
  & 57.21 & 0
  & 72.52 & 0 \\
IRCoT
  & \textbf{85.33} & 5.06
  & 53.87 & 4.38
  & \textbf{65.28} & 5.34
  & 72.80 & 5.51 \\
\cline{1-9}
DRAGIN
  & 58.33 & 3.52
  & 53.74 & 5.70
  & 58.19 & 4.16
  & 73.78 & 2.76 \\
SeaKR
  & 79.00 & 2.43
  & 54.91 & 4.06
  & 58.07 & 2.83
  & 63.88 & 2.56 \\
\cline{1-9}
ReProbe
  & 77.33 & \underline{2.41}
  & 56.19 & 3.13
  & 62.59 & 2.90
  & 72.66 & \underline{2.35} \\
\rowcolor{yellow!20}
\textsc{PUlSE}\textsubscript{step}
  & 77.00 & \textbf{2.05}
  & \underline{58.57} & \textbf{2.72}
  & \underline{65.16} & \textbf{2.52}
  & \underline{74.90} & \textbf{1.93} \\
\rowcolor{yellow!20}
\textsc{PUlSE}\textsubscript{front}
  & \underline{81.67} & 2.75
  & \textbf{58.83} & \underline{3.00}
  & 63.08 & \underline{2.61}
  & \textbf{76.43} & 2.62 \\

\hline
\multicolumn{9}{l}{\emph{Backbone: Llama 3.1 8B Instruct}} \\
\hline
CoT
  & 62.00 & 0
  & 53.47 & 0
  & 53.79 & 0
  & 60.15 & 0 \\
IRCoT
  & \textbf{87.67} & 6.96
  & \textbf{58.31} & 6.76
  & \underline{62.22} & 8.60
  & \textbf{71.69} & 8.29 \\
\cline{1-9}
DRAGIN
  & 63.67 & 3.89
  & 51.49 & 2.80
  & 54.77 & 3.00
  & 62.76 & 2.42 \\
SeaKR
  & 71.67 & 3.57
  & 55.72 & 4.94
  & 58.01 & 4.04
  & 62.34 & 2.66 \\
\cline{1-9}
ReProbe
  & 80.33 & 2.62
  & 55.66 & 2.52
  & 56.23 & 2.46
  & 67.09 & 2.98 \\
\rowcolor{yellow!20}
\textsc{PUlSE}\textsubscript{step}
  & 82.33 & \underline{2.28}
  & \underline{57.84} & \underline{2.01}
  & 60.15 & \underline{2.39}
  & 68.62 & \textbf{1.78} \\
\rowcolor{yellow!20}
\textsc{PUlSE}\textsubscript{front}
  & \underline{83.67} & \textbf{2.05}
  & 57.78 & \textbf{1.86}
  & \textbf{63.50} & \textbf{2.36}
  & \underline{70.57} & \underline{2.34} \\
\hline
\end{tabular}
\end{table*}
\paragraph{Datasets.}
We evaluate on questions entirely \emph{unseen} during probe training: 
(1)~MedMCQA~\citep{pal2022medmcqa}, a large-scale medical entrance exam dataset covering diverse medical subjects; 
(2)~JAMA Clinical Challenge~\citep{chen-etal-2025-benchmarking}, challenging case-based medical questions from the \textit{Journal of the American Medical Association}; and 
(3)~the health and biology subsets of MMLU-Pro~\citep{wang2024mmlu}, two challenging multiple-choice benchmarks covering diverse biomedical topics.

\paragraph{Baselines.}
We compare against methods spanning the retrieval-frequency spectrum:
\textbf{CoT}~\citep{wei2022chain} (no retrieval),
\textbf{IRCoT}~\citep{trivedi-etal-2023-interleaving} (retrieval at every step),
\textbf{DRAGIN}~\citep{su-etal-2024-dragin} (output-based adaptive retrieval),
\textbf{SeaKR}~\citep{yao-etal-2025-seakr} (hidden-state-based adaptive retrieval), and
\textbf{ReProbe}~\citep{Ni2025ReProbeET} (a hidden-state probe used as a drop-in retrieval trigger within our framework).
All methods share the backbone LLM, retriever, and corpus; answer generation uses greedy decoding. 
Full descriptions are in Appendix~\ref{appendix:baselines}.
    
\paragraph{External knowledge source.}

We use GPT-5.1 as an external knowledge provider to prepare factual statements offline.
The collection is prepared offline for each benchmark, with construction inputs, answer-neutralization, and quality-control procedures detailed in Appendix~\ref{appendix:construction}.
All methods access the same fixed knowledge collection through BM25, enabling comparison of retrieval-triggering strategies with a shared external knowledge source.
We additionally evaluate on independent public medical corpora to examine performance across different knowledge sources and retrieval configurations (Appendix~\ref{appendix:medrag_pilot}).

\paragraph{Evaluation metrics.}
We report answer accuracy (\%) as the primary metric and the average number of retrieval calls per instance (\#Ret) to quantify retrieval efficiency.
\begin{figure}[t]
  \centering
  \includegraphics[width=\columnwidth]{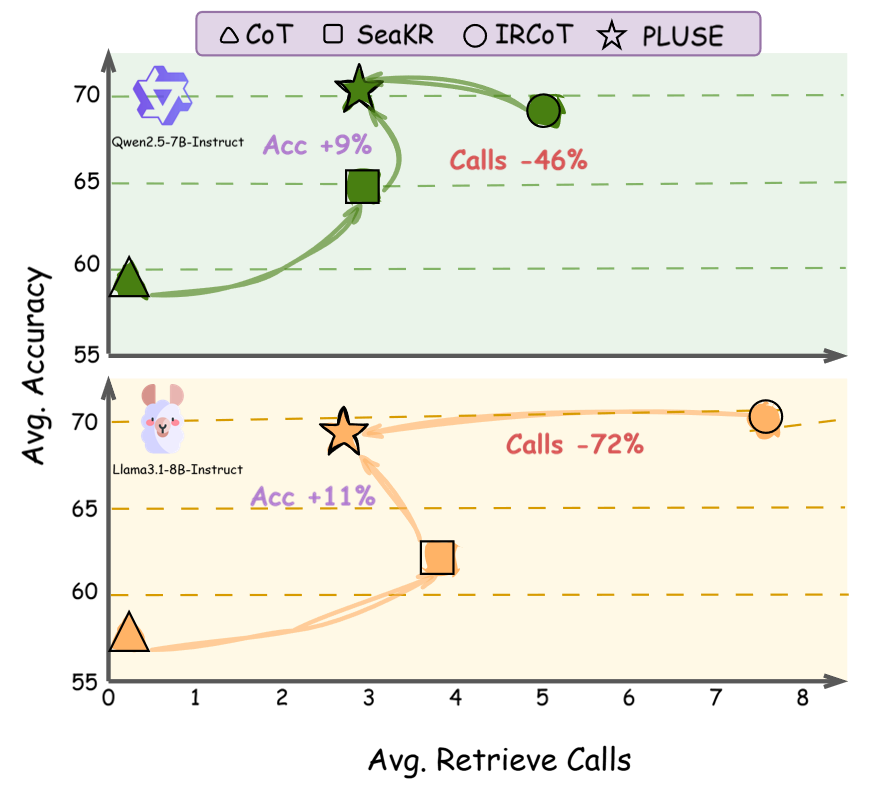}
  \caption{Accuracy--efficiency Pareto front on the four medical QA benchmarks (averaged). 
  \textbf{Top}: Qwen2.5-7B-Instruct; \textbf{Bottom}: Llama3.1-8B-Instruct. 
  The $x$-axis reports the mean number of retrieval calls per question and the $y$-axis reports average accuracy. }
  \label{fig:pareto}
\end{figure}
\paragraph{Results and analysis.}

Table~\ref{tab:main} presents end-to-end results on four medical QA
benchmarks for both injection strategies;
Figure~\ref{fig:pareto} visualizes the accuracy--efficiency trade-off.

\emph{Supplying external evidence at detected steps yields large accuracy gains.}
Compared with unassisted CoT, \textsc{PUlSE} produces substantial improvements across all benchmarks and backbones (e.g., +24.67\% on MedMCQA and +8.53\% on JAMA with Qwen; +21.67\% and +10.42\% on MedMCQA and Biology with Llama).
The gains from evidence-conditioned regeneration at probe-flagged steps provide additional support for the usefulness of the detection signal beyond the correlation-based metrics in Table~\ref{tab:probing}. On JAMA, both variants outperform IRCoT and ReProbe in paired tests (Holm-adjusted $p<0.05$; Appendix~\ref{appendix:paired_statistics}).

\emph{More precise targeting achieves greater gains with fewer retrievals.}
DRAGIN and SeaKR, despite moderate retrieval frequency, often fail to improve---and sometimes degrade---accuracy relative to CoT (e.g., DRAGIN drops to 51.49\% on JAMA with Llama versus 53.47\% for CoT), suggesting that imprecise triggering introduces noise rather than useful knowledge.
Under the same step-level injection protocol, \textsc{PUlSE}\textsubscript{step} consistently outperforms ReProbe with fewer calls: on Qwen, +2.38\% on JAMA with 13\% fewer calls, +2.57\% on Health with 13\% fewer calls, and +2.24\% on Biology with 18\% fewer calls; similar patterns hold on Llama (+2.18\% on JAMA with 20\% fewer calls, +1.53\% on Biology with 40\% fewer calls).
This simultaneous improvement on both axes confirms that trajectory geometry provides a more precise error signal than concatenated hidden states.

\emph{Targeted retrieval matches exhaustive retrieval with fewer calls.} 
IRCoT retrieves at every step (5--9 calls per instance). \textsc{PUlSE}\textsubscript{front} surpasses it on JAMA (+4.96\%) and MMLU-Pro (Biology) (+3.63\%) with Qwen using only 2.6--3.0 calls, and on MMLU-Pro (Health) (+1.28\%) with Llama at 2.36 versus 8.60 calls. On the remaining benchmarks it stays within 2--4\% at roughly one-third the retrieval budget. Retrieving for the right steps matters more than retrieving for all steps. 
Both injection variants improve over unassisted CoT across all
benchmarks and backbones while using fewer retrieval calls than
IRCoT, with their relative accuracy varying across settings.
These results support the utility of layer-wise trajectory features
for guiding selective access to external knowledge.

\section{Analysis}

This section addresses two questions: whether the probe's design choices (feature type and layer interval) are well-justified (Section~\ref{sec:design}), and whether the detected signal generalizes beyond the training domain (Section~\ref{sec:cross}).

\subsection{Ablation}
\label{sec:design}

\paragraph{Feature ablation.}
Table~\ref{tab:ablation} compares detection and downstream performance through component ablations of the fixed feature design on MMLU-Pro (Biology), with the probe trained on MedMCQA (out-of-distribution). 
Using magnitude alone yields the lowest AUROC (0.6565) and under-triggers retrieval (2.51 calls), leaving many factual errors undetected.
Using angle alone improves detection (AUROC 0.6866) but over-triggers (2.94 calls), introducing unnecessary retrievals that may distract the reasoning process.
Combining both features achieves the best AUROC (0.7273) and downstream accuracy (76.43\%) at moderate retrieval cost (2.62 calls), confirming that the two dimensions capture complementary aspects of the model's internal state and are both necessary for reliable factual error detection.

\begin{table}[t]
\centering
\caption{Feature ablation on MMLU-Pro (Biology) with Qwen2.5-7B-Instruct. Probe is trained on MedMCQA and evaluated out-of-distribution on MMLU-Pro (Biology).}
\label{tab:ablation}
\renewcommand{\arraystretch}{1}
\setlength{\tabcolsep}{2pt}
{
\begin{tabular}{l cc cc}
\hline
& \multicolumn{2}{c}{\textbf{Detection}} & \multicolumn{2}{c}{\textbf{End-to-End}} \\
\cline{2-3} \cline{4-5}
\textbf{Feature} & AUROC & PR AUC & Acc(\%) & Avg \#Ret \\
\hline
Mag       & .6565 & .3547 & 75.31 & 2.51 \\
Angle     & .6866 & .3903 & 75.59 & 2.94 \\
Mag + Ang & \textbf{.7273} & \textbf{.4316} & \textbf{76.43} & 2.62 \\
\hline
\end{tabular}
}
\end{table}

\begin{figure}[t]
    \centering
    \includegraphics[width=\columnwidth]{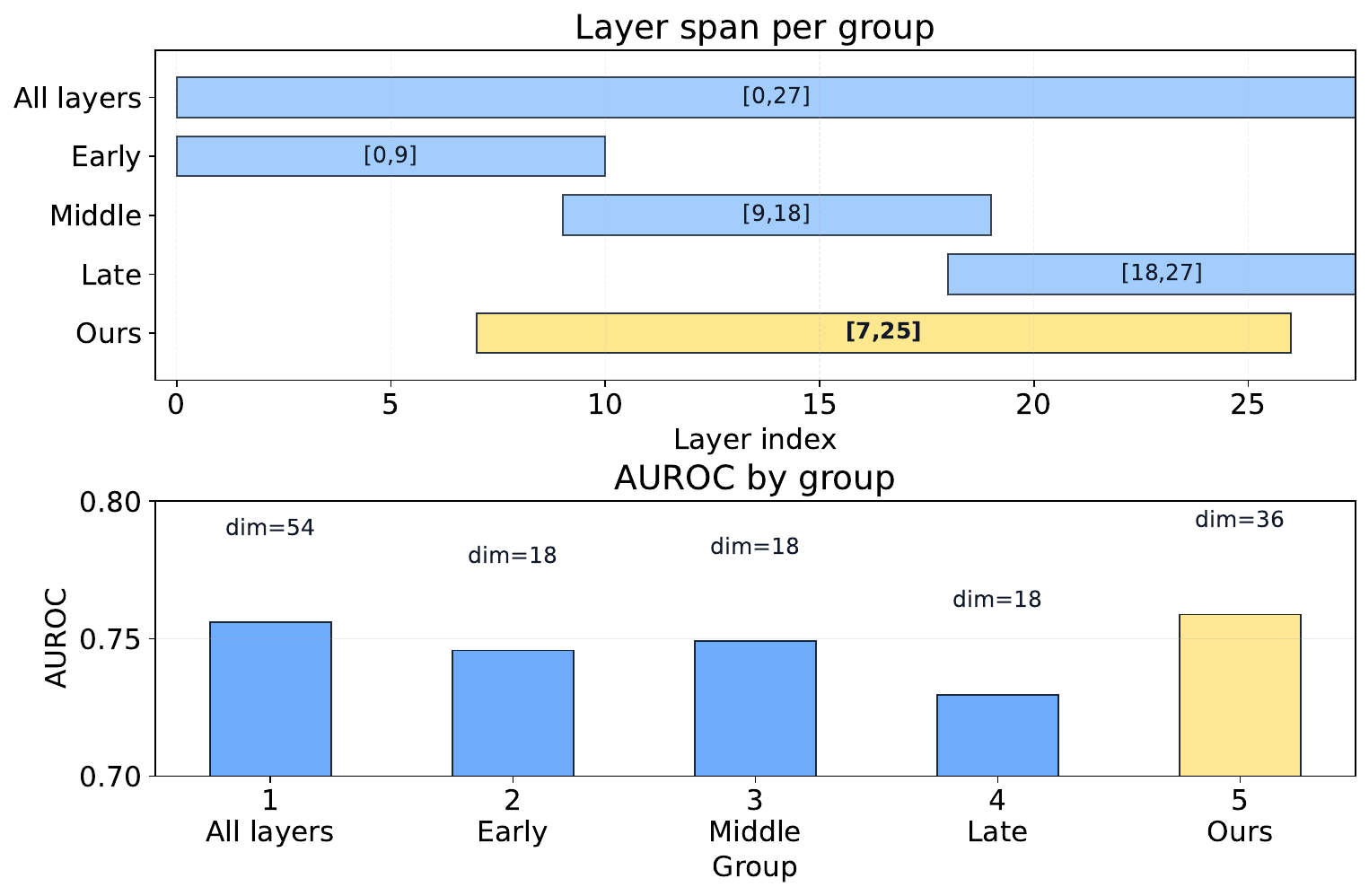}
    \caption{Layer-interval AUROC on the MedMCQA training-validation split (Qwen2.5-7B-Instruct).} 
    \label{fig:layer}
\end{figure}

\paragraph{Layer interval selection.}
Figure~\ref{fig:layer} compares layer intervals on the MedMCQA training-validation split. 
The late-only interval ([18,27]) performs worst (AUROC 0.7295): when restricted to final layers, the representations increasingly commit to the output distribution and lose discriminative signal about factual errors.
The early interval ([0,9]) also underperforms (AUROC 0.7457), as shallow layers primarily encode token-level syntax before higher-level reasoning representations have formed.
The middle interval ([9,18]) reaches 0.7491, confirming that intermediate layers carry the strongest per-segment signal.

Our selected interval [7,25] achieves the highest AUROC (0.7588) and F1 (0.5474), outperforming even the all-layer configuration (AUROC 0.7559, F1 0.5378) despite using only 36 of 54 features. This indicates that early and late layers contribute domain-specific token-distribution noise that harms both discrimination and generalization; excluding them yields a more robust and compact representation. We therefore adopt [7,25] as the default interval. Llama 3.1 exhibits the same pattern; we adopt layers [6,31] following the same selection logic. Since retrieval is triggered by the probe's binary output, this
choice directly determines when evidence is supplied during reasoning.

\subsection{Cross-Domain Generalization}
\label{sec:cross}

To test whether the geometric signal is domain-agnostic, we directly apply the probe trained exclusively on medical data (MedMCQA + AfriMedQA) to the MMLU-Pro Law subset (1{,}101 instances) without any adaptation.
As shown in Table~\ref{tab:cross_domain}, \textsc{PUlSE} achieves 42.23\% accuracy with 2.36 retrieval calls per instance, outperforming both the CoT baseline (28.52\%) and IRCoT (39.96\% at 4.48 calls).
The probe attains higher accuracy than exhaustive retrieval with roughly half as many retrieval calls, supporting transfer of triggering decisions to an unseen domain. 
Combined with the layer-selection analysis above, these results support the interpretation that middle-layer trajectory fluctuations reflect a general computational signature of failed internal knowledge retrieval, independent of the specific domain of the missing knowledge.

\begin{table}[t]
\centering
\caption{Cross-domain transfer to MMLU-Pro (Law) with Qwen2.5-7B-Instruct. The probe is trained solely on medical data and applied without adaptation.}
\label{tab:cross_domain}
\renewcommand{\arraystretch}{1}
\begin{tabular}{l c c}
\hline
\textbf{Method} & \textbf{Acc (\%)} & \textbf{Avg \#Ret} \\
\hline
CoT        & 28.52 & 0 \\
IRCoT      & 39.96 & 4.48 \\
\textsc{PUlSE}       & \textbf{42.23} & \textbf{2.36} \\
\hline
\end{tabular}
\end{table}
\section{Conclusion}
We have shown that factually erroneous reasoning steps exhibit a
distinctive geometric signature in middle-layer hidden states—a
latent \emph{pulse} encoded in the model's forward pass.
By monitoring this signal at the step level, \textsc{PUlSE} grounds
retrieval timing in the model's computational trajectory,
matching or surpassing exhaustive per-step retrieval with only
28--54\% of the calls across four medical benchmarks and transferring
zero-shot to legal reasoning.
Just as a physician takes a patient's pulse to decide when to intervene, a reasoning system can take the pulse of its hidden dynamics to decide when external knowledge is needed.
Future work may extend trajectory-based error signals as process-level
rewards in agentic settings and verify whether the signature persists
at larger model scales.

\clearpage 
\section*{Limitations}

\paragraph{Model access.}
\textsc{PUlSE} requires access to intermediate hidden states, limiting its applicability to open-weight models. It cannot be directly deployed on black-box APIs.

\paragraph{Step structure.}
The method assumes the model produces explicit step-by-step reasoning with clear boundaries (e.g., special token-delimited steps). For free-form or single-paragraph generations, more flexible logical segmentation strategies—such as semantic chunking or clause-level boundary detection—may be needed to define the units over which trajectory features are computed.
\bibliography{pulse}
\clearpage 
\appendix
\section{Dataset and Annotation Details}
\label{appendix:data}

\subsection{Data Sources and Splits}
\label{appendix:data_sources}

Our experimental pipeline involves three categories of data: (1) probe training data, used to train the factual error detection probe; (2) probe evaluation data, used to assess detection performance under both in-distribution (ID) and out-of-distribution (OOD) settings; and (3) end-to-end evaluation data, used to measure downstream QA accuracy when the probe is integrated with retrieval-augmented generation.

For probe training, we combine MedMCQA~\citep{pal2022medmcqa} and AfriMedQA~\citep{nimo-etal-2025-afrimed} to ensure diversity in question difficulty and cultural context. For probe evaluation, we hold out a separate MedMCQA subset as the ID test and use MMLU-Pro-Biology~\citep{wang2024mmlu} as the OOD test, which differs in domain granularity and question style.
Table~\ref{tab:question_stats} summarizes question-level statistics, and Table~\ref{tab:step_stats} provides step-level statistics for the probe training and evaluation splits.

\begin{table}[h]

\centering
\caption{Question-level statistics across all datasets.}
\label{tab:question_stats}
\setlength{\tabcolsep}{2pt}
\begin{tabular}{lccc}
\hline
\textbf{Role} & \textbf{Questions} & \textbf{Avg Steps} & \textbf{Domain} \\
\hline
Train/Valid & 1173 &3.7  & Medicine \\
Test (ID) & 300 & 4.9 & Medicine \\
Test (OOD) & 717 & 5.9 & Biology\\ 
\hline
\end{tabular}
\end{table}

\begin{table}[h]
\centering
\caption{Step-level statistics for probe training and evaluation. Error Ratio denotes the proportion of steps labeled as containing a factual error.}
\label{tab:step_stats}
\setlength{\tabcolsep}{2pt}
\begin{tabular}{lcccc}
\hline
\textbf{Split} & \textbf{ Total} & \textbf{No-Error} & \textbf{Error} & \textbf{Error Ratio} \\
\hline
Train/Valid& 4348 & 3216 & 1132 & 26.0\% \\
Test (ID) & 1483 & 944 & 539 & 36.4\% \\
Test (OOD) & 4257 & 3314 & 943 & 22.1\% \\
\hline
\end{tabular}
\end{table}

\subsection{Step Segmentation Protocol}
\label{appendix:step_seg}

To obtain discrete reasoning steps from chain-of-thought outputs, we instruct the backbone model to structure its response using explicit step prefixes. The generation prompt includes the following instruction:
\begin{tcolorbox}[colback=gray!10,colframe=black!50]
\texttt{Think step by step and output the reasoning process before answering. 
Each step should be started with "Step".}
\end{tcolorbox}

We then segment the generated output using rule-based matching: each segment beginning with the pattern \texttt{"Step [N]:"} is extracted as an individual reasoning step. Steps that do not conform to this format (e.g., malformed outputs or steps lacking the required prefix) are discarded. In practice, the vast majority of generated outputs conform to the prescribed format across all datasets.

Figure~\ref{fig:annotation_example} provides a complete example of a segmented reasoning chain along with its step-level annotations.

\begin{figure*}[h]
\centering
\begin{tcolorbox}[colback=gray!5, colframe=gray!60, title=Annotated Example, fonttitle=\bfseries\small, fontlower=\small]
\small
\textbf{Question:} Characteristic of venous blood flow of lower limb in duplex Doppler is? \\
\textbf{Options:} (A) Monophasic \quad (B) Biphasic \quad (C) Triphasic \quad (D) Non phasic \\
\textbf{Ground Truth:} A (Monophasic) \quad \textbf{Model Answer:} C (Triphasic) 

\tcblower
\textbf{Step 1:} Understanding the context of the question, we need to consider the nature of venous blood flow in the lower limbs and how it is visualized using duplex Doppler ultrasound. \\
\hfill \colorbox{green!15}{\texttt{Label: No Error}} \\[4pt]

\textbf{Step 2:} Duplex Doppler ultrasound is a technique that combines the Doppler effect with B-mode imaging to visualize blood flow and its direction. \\
\hfill \colorbox{green!15}{\texttt{Label: No Error}} \\[4pt]

\textbf{Step 3:} In the lower limbs, venous blood flow typically shows a characteristic pattern due to the presence of valves that prevent backflow. \\
\hfill \colorbox{green!15}{\texttt{Label: No Error}} \\[4pt]

\textbf{Step 4:} When blood flows through the veins, it can be influenced by factors such as muscle contraction, gravity, and the presence of valves. This results in a flow pattern that can be categorized into phases. \\
\hfill \colorbox{green!15}{\texttt{Label: No Error}} \\[4pt]

\textbf{Step 5:} The most common pattern observed in the lower limb veins using duplex Doppler is a triphasic flow, which includes a systolic phase, a diastolic phase, and a post-contraction phase. \\
\hfill \colorbox{red!15}{\texttt{Label: Factual Error}} \\
\textit{Reason: Incorrectly identifies lower limb venous flow as triphasic. Venous flow is monophasic, modulated by respiration and muscular compression. Triphasic flow is characteristic of peripheral arterial waveforms.} \\[4pt]

\textbf{Step 6:} This triphasic pattern is due to the sequential opening and closing of the venous valves, which allows blood to flow in one direction despite changes in pressure. \\
\hfill \colorbox{red!15}{\texttt{Label: Factual Error}} \\
\textit{Reason: Fabricates a nonexistent physiological mechanism; sequential valve opening/closing does not produce triphasic flow in veins.} \\[4pt]

\textbf{Step 7:} Considering the options provided, the correct answer is the one that matches the triphasic pattern observed in the lower limb veins. \\
\hfill \colorbox{red!15}{\texttt{Label: Factual Error}} \\
\textit{Reason: Asserts that triphasic flow is observed in lower limb veins, which is factually incorrect.}
\end{tcolorbox}
\caption{A complete annotated example showing the question, segmented reasoning steps, and step-level factual error labels with justifications.}
\label{fig:annotation_example}
\end{figure*}

\subsection{Factual Error Annotation Protocol}
\label{appendix:annotation_protocol}

\paragraph{Definition.}
A reasoning step is labeled as a \textit{factual error} if at least one atomic factual claim asserted in that step is incorrect or unsupported by established domain knowledge. Each step is evaluated \textbf{independently}. Repeating or reasserting an earlier false factual claim remains an error in the current step. In contrast, a valid conditional inference from an erroneous premise is not labeled erroneous solely because that premise was introduced earlier or the conclusion is inapplicable to the question; the current step must itself assert an incorrect factual claim. Thus, inherited error refers to a false claim repeated in the step, not automatic propagation of a preceding step's label. 

In Figure~\ref{fig:annotation_example}, Steps~6--7 are positive because they assert, respectively, a false physiological mechanism and the false claim that triphasic flow is observed in lower-limb veins. Their labels do not follow merely from preceding errors. 

\paragraph{Automated Annotation with GPT.}
We employ GPT to generate step-level annotations at scale. The annotator model receives the question, the correct answer, and the model's full reasoning chain, and is instructed to evaluate each step independently. The complete prompt template is shown below:

Using this protocol, we annotated 4,348 steps in the training set and 5,740 steps across the two probe evaluation sets.

\paragraph{Human Verification.}
Two graduate students divided the questions for manual verification of GPT-generated step labels. Each reviewer decomposed the assigned steps into atomic factual claims, retrieved medical or biological references, and checked the claims against those sources. All chains with an incorrect final answer were reviewed, and identified labeling errors were corrected. For correct-answer chains, an additional random sample of 100 questions (approximately 400 steps) was reviewed, with step-level labeling accuracy above 95\%. This was a division-of-work verification procedure rather than independent double annotation. 

The review assessed reference-supported factual consistency of individual atomic claims. The original annotation prompt below evaluates each step independently, consistent with distinguishing repeated false claims from valid conditional deductions. 
\begin{figure*}
\begin{tcolorbox}[colback=gray!5, colframe=gray!60, title=Annotation Prompt Template, fonttitle=\bfseries\small]
\small
\texttt{You are a teacher specializing in identifying errors in students' problem-solving processes. You will be provided with a question, the standard answer, and a student's response containing reasoning steps numbered by "Step". Please provide your evaluation entirely in English and strictly follow the structure below:}

\texttt{1. Step Evaluation} \\
\texttt{Evaluate each "Step" in the student's response independently. For each step, use the following fixed format:}

\texttt{\textbf{Step [number]}: [Correct / Incorrect]} \\
\texttt{\quad \textbf{Reason}: [For an incorrect step, briefly explain the reason. The evaluation should be based only on whether the step itself contains a knowledge error; mark a step as Incorrect only if the step, when judged independently, contains a knowledge error. Do not mark a step as Incorrect merely because of an error in a previous step, an error in a later step, an incorrect final answer, or inconsistency with the standard answer.]}

\texttt{2. Final Label} \\
\texttt{[Output a Python dictionary-formatted string, where the keys are step numbers and the values are correctness labels.]} \\
\texttt{\{'Step 1': 'Correct', 'Step 2': 'Incorrect', ...\}}
\end{tcolorbox}
\caption{Annotation Prompt Template}
\end{figure*}
\subsection{Class Distribution and Training Considerations}
\label{appendix:class_dist}

As shown in Table~\ref{tab:step_stats}, the training set exhibits a class imbalance with an error ratio of 26.0\% (approximately 1:2.8 positive-to-negative ratio). To mitigate the effect of this imbalance on probe training, we apply a positive class weight of 3 in the binary cross-entropy loss, which approximately matches the inverse class frequency ratio. This weighting ensures that the probe does not trivially learn to predict the majority class. We found results to be robust across weight values in the range [2, 5].

The error ratio varies across evaluation sets: the ID test shows a higher error ratio (36.4\%) because it is drawn from questions the model frequently answers incorrectly, while the OOD test has a lower error ratio (22.1\%) reflecting a different difficulty distribution. This variation allows us to assess probe robustness under different prevalence conditions, which is why we report both AUROC (threshold-invariant) and PR AUC (sensitive to class prevalence) as evaluation metrics.
\section{Baseline Details}
\label{appendix:baselines}

\subsection{Baselines for Factual Error Detection (Section~\ref{sec:probe_exp})}

\paragraph{Perplexity (PPL)~\citep{si2022prompting}.}
For each reasoning step, we compute the average token-level negative log-likelihood (i.e., surprisal) under the backbone model. A higher perplexity indicates that the model assigns lower probability to its own generation, which may signal factual uncertainty. We treat perplexity as a step-level anomaly score and select the detection threshold via Youden's J statistic on the validation set.

\paragraph{MaxProb.}
For each reasoning step, we record the maximum softmax probability across all generated tokens and average these values as the step-level confidence score. Lower average maximum probability suggests the model is less committed to its token choices, potentially indicating erroneous content. The decision threshold is again selected via Youden's J statistic.

\paragraph{Entropy~\citep{kuhn2023semantic}.}
We compute the Shannon entropy of the next-token probability distribution at each decoding position, then average over all tokens in the step and normalize by sequence length. Higher entropy reflects greater distributional uncertainty. The threshold is selected identically to the other output-level methods.

\paragraph{CoE-R / CoE-C~\citep{wang2025latent}.}
Both methods compute the magnitude change and angular change between
adjacent hidden states across all layers for a reasoning step.
CoE-R averages the difference between local magnitude
and angular ratios into a single scalar; CoE-C treats each layer
transition as a complex number and takes the magnitude of their
average resultant vector. Both are training-free and compress the
full layer-wise progression into one score per step.

\paragraph{ReProbe~\citep{Ni2025ReProbeET}.}
ReProbe concatenates the hidden-state vectors from all layers at the final token position of a reasoning step, forming a high-dimensional representation of the step's full internal state. This concatenated vector is then projected and classified by a lightweight Transformer encoder trained with binary cross-entropy. Unlike our method, ReProbe does not explicitly model the sequential, layer-by-layer structure of the representation trajectory.
\subsection{Baselines for Retrieval Experiments (Section~\ref{sec:retrieval_expe})}

\paragraph{CoT~\citep{wei2022chain}.}
Standard chain-of-thought prompting without any retrieval, representing
the backbone model's intrinsic reasoning capability.

\paragraph{IRCoT~\citep{trivedi-etal-2023-interleaving}.}
At every reasoning step, the system issues a retrieval call, prepends
the top-$k$ passages to the query, and continues generation.
This represents an indiscriminate full-retrieval strategy and serves
as an upper bound on retrieval frequency.

\paragraph{DRAGIN~\citep{su-etal-2024-dragin}.}
Retrieval is triggered when the maximum token-level entropy within a
reasoning step exceeds a calibrated threshold. The retrieval query is
constructed from the tokens with the highest attention-weighted
information gain in the flagged step.

\paragraph{SeaKR~\citep{yao-etal-2025-seakr}.}
The system samples multiple continuations for the current step and
estimates the model's internal uncertainty via hidden-state covariance
across samples. Retrieval is invoked when this covariance exceeds a
threshold, indicating that the model lacks sufficient parametric
support. A self-refinement module then selects between the original
and the retrieval-augmented continuation.

\paragraph{ReProbe~\citep{Ni2025ReProbeET}.}
A lightweight Transformer probe is trained on the concatenation of
hidden states across all layers at the final token of each step.
We integrate its binary classification output into our retrieval
framework as a drop-in trigger replacement: retrieval is called
whenever ReProbe flags a step as factually erroneous, using the same
injection strategy and corpus as \textsc{PUlSE}.

\subsection{Robustness to Explicit Uncertainty Expressions}
\label{appendix:non_hedging}

We evaluate the existing Qwen2.5-7B-Instruct reasoning traces with frozen PULSE and ReProbe weights, without regenerating reasoning or retraining either probe. We retain steps with aligned reasoning text, cached representations, and labels. The resulting diagnostic pools contain 1,464 MedMCQA steps and 4,097 MMLU-Pro (Biology) steps; these aligned pools are evaluated separately from the main detection table. 

We exclude an entire step if it contains a case-insensitive, word-boundary match to a predefined uncertainty lexicon: \emph{may, might, could, unclear, uncertain, not clear, not immediately clear, not well established, possibly, perhaps}, and \emph{suggest/suggests/suggested/suggesting}. The original step text and representations are otherwise unchanged. Both probes are evaluated on identical subsets. 

\begin{table*}[t]
\centering
\small
\caption{Detection before and after filtering explicit uncertainty expressions. ``All aligned'' is the diagnostic pool before filtering; ``Non-hedging'' excludes lexicon-matched steps. PR-AUC uses trapezoidal integration. Error prevalence is reported because filtering changes class balance.} 
\label{tab:non_hedging}
\begin{tabular}{llrrrrrr}
\toprule
& & & & \multicolumn{2}{c}{PULSE} & \multicolumn{2}{c}{ReProbe} \\
\cmidrule(lr){5-6}\cmidrule(lr){7-8}
Dataset & Subset & Steps & Error (\%) & AUROC & PR-AUC & AUROC & PR-AUC \\
\midrule
MedMCQA & All aligned & 1,464 & 36.3 & .746 & .583 & .728 & .588 \\
& Non-hedging & 1,365 & 35.1 & .744 & .570 & .732 & .575 \\
MMLU-Pro (Biology) & All aligned & 4,097 & 22.4 & .712 & .417 & .694 & .438 \\
& Non-hedging & 3,662 & 21.2 & .704 & .386 & .697 & .431 \\
MedBullets$^{\dagger}$ & All aligned & 844 & 35.4 & .752 & .564 & .736 & .609 \\
& Non-hedging & 571 & 32.6 & .742 & .524 & .749 & .598 \\
\bottomrule
\end{tabular}
\end{table*}

PULSE retains discrimination on the non-hedging subsets (Table~\ref{tab:non_hedging}). Its AUROC changes from .746 to .744 on MedMCQA and from .712 to .704 on Biology. ReProbe also retains predictive performance and attains higher PR-AUC on these subsets. These results indicate that PULSE's signal is not confined to steps containing the listed explicit uncertainty expressions. 

We compute 95\% percentile confidence intervals using 2,000 question-level bootstrap resamples, retaining all eligible steps from each sampled question (seed 20260918). PULSE's non-hedging AUROC intervals are [.714, .775] on MedMCQA and [.677, .728] on Biology. 

\paragraph{Auxiliary MedBullets analysis.}
The MedBullets rows marked $\dagger$ cover 198 questions after excluding 92 records with inconsistent step counts, 16 conflicting annotations, and two duplicate traces from 308 cached records. They retain the historical convention of treating missing evaluation entries as Correct; this auxiliary result is distinguished from the primary MedMCQA and Biology diagnostics. 

\subsection{Paired Statistical Analysis on JAMA}
\label{appendix:paired_statistics}

We align the saved Qwen2.5-7B-Instruct predictions by question text and complete answer options across all 1,511 JAMA questions. For each comparison, we compute the accuracy difference using 20,000 paired question-level bootstrap resamples (seed 20261011) and report its 95\% percentile interval. We additionally apply a two-sided exact McNemar test and Holm correction across the four PULSE--baseline comparisons in Table~\ref{tab:jama_paired}. These intervals describe question-level uncertainty for the fixed evaluated runs. 

\begin{table*}[t]
\centering
\small
\caption{Paired JAMA comparisons ($n=1,511$), retaining the main table's answer-scoring convention. Differences and confidence intervals are in percentage points. Wins/losses count questions answered correctly only by PULSE or only by the baseline. All four comparisons remain significant after Holm correction within this family.} 
\label{tab:jama_paired}
\begin{tabular}{llrrrrr}
\toprule
PULSE variant & Baseline & Difference & 95\% CI & Wins/losses & Exact $p$ & Holm $p$ \\
\midrule
Step & IRCoT & +4.70 & [2.38, 7.15] & 209/138 & .000163 & .000490 \\
Step & ReProbe & +2.38 & [0.40, 4.43] & 140/104 & .024849 & .024849 \\
Front & IRCoT & +4.96 & [2.65, 7.35] & 207/132 & .000055 & .000219 \\
Front & ReProbe & +2.65 & [0.66, 4.63] & 139/99 & .011315 & .022630 \\
\bottomrule
\end{tabular}
\end{table*}

\paragraph{Answer-extraction sensitivity.}
The main-table scoring counts are 885 (PULSE-step), 889 (PULSE-front), 814 (IRCoT), and 849 (ReProbe). Requiring an explicit final \texttt{Answer:} option label changes one ReProbe outcome, reducing its count to 848; the other counts are unchanged. Under this strict extraction, PULSE-step and PULSE-front exceed ReProbe by 2.45 and 2.71 points, with intervals [0.46, 4.50] and [0.73, 4.70], respectively. All four Holm-adjusted tests remain below .05. Differences are computed from these counts, before percentage rounding; 889/1,511 is 58.8352\%, while the main table retains its originally reported 58.83\%. 

\section{External Knowledge Sources and Retrieval Setup}
  \label{appendix:corpus}

  This appendix examines how corpus coverage and retrieval configuration
  interact with retrieval timing. We first evaluate retrieval-augmented
  reasoning using benchmark-independent public medical corpora
  (Section~\ref{appendix:medrag_pilot}). The first experiment presents a
  failure case caused by insufficient corpus support, while the second
  examines whether selective retrieval remains effective when the public
  corpus is better aligned with the benchmark. We then describe how factual knowledge is obtained from GPT-5.1 and prepared for the main experiments
(Section~\ref{appendix:construction}), followed by the retrieval
configuration (Section~\ref{appendix:retrieval_config}).

  \subsection{Evaluation with Public Medical Corpora}
  \label{appendix:medrag_pilot}

  \paragraph{JAMA with the MedRAG Textbook corpus.}
  To assess factual-error-triggered retrieval under realistic conditions,
  we first conducted a pilot experiment using the open-source Textbook
  corpus from MedRAG~\citep{medrag}, which contains approximately 160K
  passages drawn from medical textbooks. We evaluated a random sample of
  500 JAMA questions using BM25 retrieval with the top-3 candidates,
  selecting the first non-duplicate passage at each retrieval step.

  Table~\ref{tab:medrag_pilot} reports the results. All three
  retrieval-augmented methods perform below the no-retrieval CoT baseline
  in this setting. \textsc{PUlSE}, which retrieves more conservatively,
  experiences a smaller degradation than IRCoT and SeaKR. More generally,
  the results indicate that repeatedly retrieving from an insufficiently
  matched corpus does not compensate for missing evidence and may instead
  expose the model to additional distractors.

  \begin{table}[h]
  \centering
  \caption{Accuracy (\%) on 500 JAMA questions using the public MedRAG
  Textbook corpus. All retrieval-augmented methods perform below the
  no-retrieval CoT baseline in this corpus--benchmark setting.}
  \label{tab:medrag_pilot}
  \begin{tabular}{lcc}
  \hline
  \textbf{Method} & \textbf{Acc. (\%)} & \textbf{Avg. Ret.} \\
  \hline
  CoT (no retrieval)    & 51.40 & 0.00 \\
  \textsc{PUlSE} (Ours) & 47.60 & 2.25 \\
  IRCoT                 & 45.40 & 3.85 \\
  SeaKR                 & 42.40 & 3.43 \\
  \hline
  \end{tabular}
  \end{table}

  We attribute this result to two compounding factors:
  (1)~\textit{insufficient coverage}, because a general textbook corpus
  does not reliably contain the fine-grained factual knowledge required
  by every JAMA reasoning step; and (2)~\textit{retrieval noise}, because
  passages that are medically related to a question may nevertheless be
  irrelevant to the specific uncertainty encountered during reasoning.
  Consequently, an appropriate retrieval trigger cannot recover knowledge
  that is absent from the corpus, while an irrelevant passage may still
  mislead the model even when retrieval occurs at a reasonable step.

  \paragraph{MedQA with the public StatPearls corpus.}
  To examine whether the preceding failure primarily reflects
  corpus--benchmark mismatch, we conduct a follow-up experiment on a reserved sample of 600
  MedQA~\cite{jin2021disease} questions. These questions are disjoint from the data used to 
  develop and freeze the final retrieval configuration.

  We use the public StatPearls corpus distributed through
  MedRAG~\citep{medrag}. The corpus is independent of the MedQA questions,
  gold labels, and generated reasoning trajectories. Passages are
  retrieved using the MedCPT dense index with a top-6 candidate budget.
  We preserve both section titles and passage contents so that the text
  presented to the model remains consistent with the representation used
  to construct the index.

  The retrieval pipeline uses the original question, its terminal task,
  and the current question-grounded reasoning focus to formulate retrieval
  queries. Retrieved evidence is presented with a neutral instruction
  stating that it may be incomplete or irrelevant and should be ignored
  when it does not directly apply.

  All methods are evaluated on the same 600 questions using the same
  Qwen2.5-7B-Instruct checkpoint, StatPearls corpus, MedCPT index,
  candidate budget, and answer-evaluation procedure. Each method retains
  its retrieval controller and, where applicable, its native
  evidence-selection mechanism. For \textsc{PULSE} and IRCoT, the same answer-blind
candidate-selection policy is used to select at most one directly relevant passage from the retrieved
  candidates. This policy observes the question and all answer choices
  symmetrically but never observes the gold label.

  Method-specific triggering and selection configurations are fixed
  before evaluation and are not tuned on the reserved 600-question
  sample. Thus, the differences in retrieval frequency arise from the
  corresponding retrieval strategies.

  \begin{table}[h]
  \centering
  \caption{Results on the reserved MedQA sample using the public
  StatPearls corpus ($n=600$). Avg.\ Ret.\ denotes the average number of
  retrieval calls per question.}
  \label{tab:medqa_public_corpus}
  \begin{tabular}{lcc}
  \hline
  \textbf{Method} & \textbf{Acc. (\%)} & \textbf{Avg. Ret.} \\
  \hline
  CoT (no retrieval)    & 61.67 & 0.00 \\
  IRCoT                 & 61.33 & 5.57 \\
  SeaKR                 & 61.33 & 3.66 \\
  \textsc{PUlSE} (Ours) & \textbf{63.50} & 0.96 \\
  \hline
  \end{tabular}
  \end{table}

  As shown in Table~\ref{tab:medqa_public_corpus}, \textsc{PUlSE}
  achieves the highest numerical accuracy. It exceeds CoT by 1.83
  percentage points, IRCoT and SeaKR by 2.17 points. At the same time,
  \textsc{PUlSE} reduces the average number of retrieval calls by 82.8\%
  relative to IRCoT, 73.8\% relative to SeaKR.

  Retrieval frequency varies across questions: \textsc{PUlSE} skips retrieval
  entirely for 230 of the 600 questions (38.3\%) and selectively invokes
  the retriever on the remaining questions. Its retrieval calls are
  therefore concentrated on reasoning trajectories identified as more
  likely to require external factual support.

  Together, the two public-corpus experiments illustrate the interaction
  between evidence availability and retrieval timing. On JAMA, limited
  corpus coverage makes retrieval harmful on average, and conservative
  triggering mainly limits the resulting degradation. On MedQA with the
  better-aligned public StatPearls corpus and a quality-controlled
  retrieval pipeline, selective retrieval improves over no retrieval
  while requiring substantially fewer retrieval calls than the retrieval
  baselines. This contrast supports the importance of retrieval timing
  without assuming that timing alone can compensate for an unsuitable
  corpus.
\subsection{Knowledge Preparation}
\label{appendix:construction}

GPT-5.1 serves as an external knowledge provider for the main experiments.
We generate factual knowledge statements offline using benchmark-specific inputs and organize them into a searchable collection for each benchmark.
The resulting collections are fixed before evaluation and shared across retrieval methods.

\paragraph{Generation procedure.}
For each question in the dataset, we provide GPT-5.1 with the question, the gold answer, and the model-generated chain-of-thought reasoning. These inputs are used only during offline corpus construction to enumerate the factual knowledge needed to support the model's possible reasoning steps, including factual misconceptions that may arise in its chain of thought. The reference answers and reasoning traces used for offline construction are not supplied as additional inputs during evaluation; retrieval queries are formed from the reasoning steps generated at inference time. The model is prompted to evaluate each reasoning step and produce the factual knowledge required to support that step. The full prompt template is shown below:
\begin{figure*}

\begin{tcolorbox}[colback=gray!5, colframe=gray!60, title=Corpus Construction Prompt Template, fonttitle=\bfseries\small]
\small
\texttt{You are a teacher specializing in identifying student errors in problem-solving. You are provided with a problem, a standard answer, and a student's response (including thought processes numbered "Step"). Please strictly adhere to the following structure when using the output:}

\texttt{\textbf{Step Evaluation}}\\
\texttt{Please evaluate each "Step" in the student's response independently. For each step, please use the following fixed format:}

\texttt{\textbf{Step} [Number]: [Correct / Incorrect]}\\
\texttt{\textbf{Reason}: [Briefly explain the reason. Important Principle: Evaluation should be based on whether the logical derivation of this step conforms to subject-specific standards. Even if a previous step is incorrect, as long as the current step's logical derivation based on the previous result is correct, it should still be judged as Correct (unless the step itself introduces new incorrect knowledge).]}\\
\texttt{\textbf{Knowledge}: [Key Fields: Please provide the multiple original factual knowledge points required to support this step of the reasoning. Content Requirements: Only definitions, theorems, formulas, constants, subject-specific knowledge, or objective facts are allowed. Prohibited Content: Specific calculation processes, reasoning results, or final conclusions related to this problem are strictly prohibited. Do not mention the problem or reveal the answer.]}
\end{tcolorbox}
\caption{Prompt template for corpus construction.}

\end{figure*}

Each step yields 1--4 knowledge statements depending on the complexity of the reasoning involved.

\paragraph{Filtering and quality control.}
We apply a three-stage answer-neutralization and quality-control process. First, rule-based matching removes statements that contain calculation processes, problem-specific reasoning chains, or direct answers (identified via keyword patterns such as ``therefore the answer is'', ``substituting into the equation'', and similar phrases). Second, a separate LLM pass screens the remaining statements for explicit or implicit answer-revealing cues and rewrites suspicious entries as standalone, answer-neutral factual knowledge. This neutralization removes option labels, correctness judgments, question-specific conclusions, and language that explicitly endorses or eliminates an answer choice, while preserving the underlying domain fact. Third, the authors conduct a random audit of the processed statements to verify that sampled entries constitute objective factual claims (e.g., definitions, theorems, and biomedical facts) rather than problem-specific reasoning or answer-bearing conclusions. Entries that fail this audit are removed or revised. Duplicate statements within each benchmark corpus are then removed. 

\paragraph{Corpus statistics.}
Table~\ref{tab:oracle_stats} summarizes the resulting corpora. Each knowledge statement is 1--3 sentences in length.

\begin{table}[h]
\centering
\caption{Corpus statistics (number of unique knowledge statements per benchmark).}
\label{tab:oracle_stats}
\begin{tabular}{lc}
\hline
\textbf{Benchmark} & \textbf{\# Statements} \\
\hline
JAMA & 20{,}225 \\
MMLUPro\_Health & 4{,}522 \\
MMLUPro\_Biology & 4{,}262 \\
MedMCQA (test, 300 questions) & 1{,}503 \\
\hline
\end{tabular}
\end{table}
\paragraph{Random-trigger control.}
To disentangle the benefit of access to external knowledge
from the additional benefit of learned retrieval timing, we replace
only the \textsc{PULSE} trigger with an independent Bernoulli trigger
on JAMA using Qwen2.5-7B-Instruct. Each completed reasoning step
triggers retrieval with probability $p=0.55$. The corpus, BM25 retriever, query construction, evidence prepending,
regeneration of the triggering step with all preceding steps preserved,
and decoding configuration are otherwise identical to
\textsc{PUlSE}$_{\mathrm{front}}$. Random-trigger results are averaged
over three seeds; the remaining rows are reproduced from
Table~\ref{tab:main} for direct comparison.
  \begin{table}[h]
  \centering
  \caption{Diagnostic random-trigger control on JAMA with
  Qwen2.5-7B-Instruct ($n=1{,}511$). Random uses a per-step trigger
  probability of $0.55$ and is averaged over three seeds. $\dagger$ denotes
  results reproduced from Table~\ref{tab:main}.}
  \label{tab:random_trigger_jama}
  \begin{tabular}{lcc}
  \hline
  \textbf{Method} & \textbf{Acc. (\%)} $\uparrow$ & \textbf{Avg. Ret.} $\downarrow$ \\
  \hline
  CoT$^{\dagger}$                         & 50.30 & 0.00 \\
  IRCoT$^{\dagger}$                       & 53.87 & 4.38 \\
  Random ($p=0.55$)              & 55.32 & 2.81 \\
  \textsc{PUlSE}$_{\mathrm{front}}^{\dagger}$ & \textbf{58.83} & 3.00 \\
  \hline
  \end{tabular}
  \end{table}

  Random triggering can also benefit from access to the shared knowledge collection.
  For any fixed set of factual-error steps, a Bernoulli trigger with
  $p=0.55$ covers 55\% of them in expectation. When such a trigger coincides
  with an erroneous step, the shared knowledge collection and the query derived from the current reasoning step provide a reasonable opportunity to retrieve
  corrective evidence, explaining why Random improves over no-retrieval CoT.
  However, Random also misses 45\% of factual-error steps in expectation and
  spends the same trigger probability on already-correct steps. At a nearly
  identical retrieval budget, \textsc{PUlSE}$_{\mathrm{front}}$ improves over
  Random by approximately 3.5 accuracy points. This result is consistent with the learned
  trigger allocating retrievals more selectively toward correctable factual
  errors while avoiding unnecessary interventions. IRCoT performs below both
  selective strategies despite substantially more retrieval calls, further
  showing that retrieval frequency alone does not determine downstream
  accuracy. Together with the answer-neutralization pipeline above, this
  control separates the baseline benefit of evidence availability from the
  additional benefit of learned retrieval timing.

\subsection{Retrieval Configuration}
\label{appendix:retrieval_config}

We use BM25 to retrieve factual statements from the GPT-5.1-generated knowledge collection. At each reasoning step where retrieval is triggered, we construct the query by extracting content words from the current step text, retaining only tokens with part-of-speech tags in \{\textsc{noun}, \textsc{propn}, \textsc{num}, \textsc{adj}, \textsc{verb}, \textsc{adv}\}. This filtering removes function words and improves retrieval precision for factual content.

We retrieve the top-3 candidate passages and select the first passage that has not already been used in a previous retrieval step for the same question, ensuring that each triggered retrieval injects novel information. The format for integrating retrieved knowledge into the prompt is illustrated in Appendix~\ref{appendix:case_study}.
\paragraph{Design rationale.}
A shared knowledge collection and fixed retrieval configuration reduce variation in the knowledge-access pipeline when comparing retrieval-triggering strategies.
The public-corpus experiments in Section~\ref{appendix:medrag_pilot} complement this comparison by examining performance with different knowledge sources and retrieval configurations.

\subsection{Computational Resources}
\label{appendix:compute}

\textsc{PUlSE} introduces only a small trainable component. The step-level error probe is implemented as a two-layer MLP with hidden dimensions 128 and 64, and takes as input low-dimensional magnitude and angular features rather than full hidden states. As a result, the probe contains only on the order of tens of thousands of trainable parameters, which is negligible compared with the frozen backbone LLM. The backbone model is not fine-tuned.

To compute the magnitude and angular trajectory features, \textsc{PUlSE} records hidden states during generation. Storage requirements depend on the number of recorded states, layers, and the backbone's hidden dimension. Retrieval calls are triggered by the probe's step-level decisions. 
Experiments were conducted on NVIDIA A100 GPUs (80 GB).
\section{Case Study}
\label{appendix:case_study}
\begin{figure*}[t]
\centering
\small
\begin{tcolorbox}[colback=gray!5, colframe=gray!60, fonttitle=\bfseries\small, title=Case-study question from JAMA Clinical Challenge.]

\begin{tcolorbox}[colback=white, colframe=gray!40, fonttitle=\bfseries\scriptsize, title=Question]
An 8-year-old girl was referred to the vitreoretinal service for evaluation of bilateral macular lesions noted by her pediatric ophthalmologist. Bilateral lateral rectus muscle recession for exotropia was performed 4 years before this presentation. Medical history was otherwise unremarkable. [\ldots] A single, 1-cm, oval, lightly pigmented macule with irregular borders was noted on her right upper arm. Dilated fundus examination revealed eccentric, yellow-clear, preretinal lesions in the maculas of both eyes. OCT revealed hyperreflective areas on the surface of the macula with a spiculated appearance, projecting anteriorly into the vitreous. \\[4pt]
\textbf{Options:} A: Obtain a family medical history and genetic testing \quad B: Recommend pars plana vitrectomy with epiretinal membrane peel \quad C: Observe the lesions \quad D: Give an intravitreal ocriplasmin injection
\end{tcolorbox}
\end{tcolorbox}
\caption{Case-study question from JAMA Clinical Challenge.}
\end{figure*}

\begin{figure*}[t]
\centering
\small
\begin{tcolorbox}[colback=gray!5, colframe=gray!60, fonttitle=\bfseries\small, title=\textsc{PUlSE} -- Case Study (JAMA Clinical Challenge)]
\begin{tcolorbox}[colback=gray!5, colframe=gray!60, fonttitle=\bfseries\scriptsize, title=Step 1 (No Error Detected)]
Analyze the patient's clinical presentation and findings. The patient is an 8-year-old girl with bilateral macular lesions. She has a history of exotropia surgery. Visual acuity is stable at 20/40 in both eyes. There is a residual intermittent exotropia. A single, 1-cm, oval, lightly pigmented macule with irregular borders is noted on her right upper arm. Dilated fundus examination reveals eccentric, yellow-clear, preretinal lesions in the maculas of both eyes. OCT shows hyperreflective areas on the surface of the macula with a spiculated appearance, projecting anteriorly into the vitreous, and increased retinal thickness with loss of the foveal contour.
\end{tcolorbox}
\begin{tcolorbox}[colback=red!5, colframe=red!50, fonttitle=\bfseries\scriptsize, title=Step 2 Draft (\textcolor{red}{Factual Error Detected by Probe})]
Consider the differential diagnosis. The bilateral epiretinal membranes with a spiculated appearance and anterior vitreous projection on OCT are atypical for standard idiopathic epiretinal membranes. The pigmented macule on the right upper arm may represent a caf\'{e}-au-lait spot, which can be associated with phakomatoses such as Neurofibromatosis Type 1 (NF1). However, the specific significance of the spiculated OCT morphology with anterior projection in a pediatric patient is \textit{unclear}. Other possibilities include combined hamartoma of the retina and retinal pigment epithelium, or vitreomacular traction syndrome. Further clinical evaluation and imaging may be needed to narrow down the differential diagnosis.
\end{tcolorbox}
\begin{tcolorbox}[colback=blue!5, colframe=blue!50, fonttitle=\bfseries\scriptsize, title=Retrieved Knowledge (1 retrieval triggered)]
Epiretinal membranes in NF2 often have a spiculated appearance with curled edges, projecting anteriorly into the vitreous and are best visualized using optical coherence tomography.
\end{tcolorbox}
\begin{tcolorbox}[colback=gray!5, colframe=gray!60, fonttitle=\bfseries\scriptsize, title=Step 2 Regenerated (After Retrieval)]
Consider the differential diagnosis. The clinical presentation and OCT findings are consistent with Neurofibromatosis Type 2 (NF2). NF2 is characterized by bilateral vestibular schwannomas and can present with retinal abnormalities, including epiretinal membranes. The cutaneous pigmented macule may raise consideration of an underlying phakomatosis, while the spiculated epiretinal membrane morphology is more directly suggestive of NF2-related ocular findings.
\end{tcolorbox}
\begin{tcolorbox}[colback=gray!5, colframe=gray!60, fonttitle=\bfseries\scriptsize, title=Step 3 (No Error Detected)]
Determine the next steps. Given the clinical suspicion of NF2, the appropriate next step is to obtain a detailed family medical history and genetic testing to confirm the diagnosis before considering treatment options.
\end{tcolorbox}
\begin{tcolorbox}[colback=green!8, colframe=green!70, fonttitle=\bfseries\scriptsize, title=Final Answer]
\textbf{A: Obtain a family medical history and genetic testing} \hfill \textcolor{green!70!black}{Correct}
\end{tcolorbox}
\end{tcolorbox}
\caption{Case study of \textsc{PUlSE} on a JAMA Clinical Challenge question.}
\label{fig:case_pulse}
\end{figure*}

\begin{figure*}[t]
\centering
\small
\begin{tcolorbox}[colback=gray!5, colframe=gray!60, fonttitle=\bfseries\small, title=ReProbe -- Case Study (Same Question)]
\begin{tcolorbox}[colback=red!5, colframe=red!50, fonttitle=\bfseries\scriptsize, title=Step 1 Draft (\textcolor{red}{Factual Error Detected by ReProbe Trigger})]
The patient is an 8-year-old girl with bilateral macular lesions and a history of bilateral lateral rectus muscle recession for exotropia. On examination, OCT reveals atypical epiretinal membranes with a spiculated appearance projecting anteriorly into the vitreous. A single pigmented macule with irregular borders is noted on the right upper arm. The combination of ocular findings and a cutaneous lesion may suggest an underlying systemic or genetic condition, but the specific clinical significance of the spiculated epiretinal membrane morphology is \textit{not immediately clear} from the presentation alone.
\end{tcolorbox}
\begin{tcolorbox}[colback=blue!5, colframe=blue!50, fonttitle=\bfseries\scriptsize, title=Retrieved Knowledge]
Observation is an option for NF2-related epiretinal membranes if vision loss is mild, but recent studies suggest gains in vision with early pars plana vitrectomy and epiretinal membrane peel.
\end{tcolorbox}
\begin{tcolorbox}[colback=gray!5, colframe=gray!60, fonttitle=\bfseries\scriptsize, title=Step 1 Regenerated (After Retrieval)]
The patient's history of bilateral lateral rectus muscle recession for exotropia and the presence of a pigmented macule on the right upper arm suggest a possible diagnosis of Neurofibromatosis Type 2 (NF2). NF2 is associated with bilateral vestibular schwannomas and can also present with retinal abnormalities, including epiretinal membranes.
\end{tcolorbox}
\begin{tcolorbox}[colback=gray!5, colframe=gray!60, fonttitle=\bfseries\scriptsize, title=Step 2--3 (No Error Detected)]
Given the clinical presentation, including the epiretinal membranes with a spiculated appearance and anterior projection into the vitreous, the findings are consistent with NF2.

The patient's visual acuity is stable at 20/40 in both eyes, but the presence of these membranes could potentially lead to further vision loss if left untreated.
\end{tcolorbox}
\begin{tcolorbox}[colback=red!5, colframe=red!50, fonttitle=\bfseries\scriptsize, title=Step 4 Draft (\textcolor{red}{Factual Error Detected by ReProbe Trigger})]
Given the diagnosis of NF2-related epiretinal membranes, treatment options should be considered. Surgical intervention such as vitrectomy may be beneficial for improving visual outcomes, but the specific indications and timing for surgery in this context are not well established.
\end{tcolorbox}
\begin{tcolorbox}[colback=blue!5, colframe=blue!50, fonttitle=\bfseries\scriptsize, title=Retrieved Knowledge]
Pars plana vitrectomy can involve procedures such as vitreous biopsy and epiretinal membrane peeling.
\end{tcolorbox}
\begin{tcolorbox}[colback=gray!5, colframe=gray!60, fonttitle=\bfseries\scriptsize, title=Step 4 Regenerated (After Retrieval)]
Recent studies have shown that early intervention with pars plana vitrectomy and epiretinal membrane peel can improve vision outcomes in patients with NF2-related epiretinal membranes.
\end{tcolorbox}
\begin{tcolorbox}[colback=gray!5, colframe=gray!60, fonttitle=\bfseries\scriptsize, title=Step 5--6 (No Error Detected)]
Genetic testing for NF2 would be beneficial to confirm the diagnosis and to inform the patient and family about the risk of developing other NF2-related conditions.

Given the clinical findings and the potential for vision improvement, the most appropriate next step would be to recommend a pars plana vitrectomy with epiretinal membrane peel.
\end{tcolorbox}
\begin{tcolorbox}[colback=red!8, colframe=red!70, fonttitle=\bfseries\scriptsize, title=Final Answer]
\textbf{B: Recommend pars plana vitrectomy with epiretinal membrane peel} \hfill \textcolor{red!70!black}{Incorrect}
\end{tcolorbox}

\end{tcolorbox}
\caption{Case study of ReProbe on the same question.}
\label{fig:case_reprobe}
\end{figure*}

\subsection*{Analysis}

Both methods detect a factual error early in the reasoning process and enter
the iterative retrieval loop. \textsc{PUlSE} detects the error at Step~2, where
the model recognizes that the OCT morphology is atypical but fails to connect
the spiculated epiretinal membrane pattern to NF2. A single targeted retrieval
provides this missing diagnostic knowledge: epiretinal membranes in NF2 often
have a spiculated appearance and project anteriorly into the vitreous. With
this diagnostic-level evidence, the regenerated step correctly identifies NF2
as the likely underlying condition, and the model proceeds to select family
history and genetic testing as the appropriate next step before treatment
decisions are made.

ReProbe, in contrast, triggers retrieval earlier at Step~1 and retrieves
management-oriented evidence about observation versus early vitrectomy for
NF2-related epiretinal membranes. Although this evidence is factually relevant,
it is premature for the question, whose correct next step is diagnostic
confirmation rather than immediate treatment. ReProbe later triggers a second
retrieval at Step~4, which injects additional procedural knowledge about pars
plana vitrectomy and epiretinal membrane peeling. This second retrieval further
primes the model toward a treatment-oriented conclusion. As a result, even
though the model mentions genetic testing in Step~5, it ultimately prioritizes
surgical intervention in Step~6 and selects the incorrect answer.

This case highlights that retrieval timing can affect downstream reasoning, even
when the retrieved evidence is factually relevant. Evidence about management or
surgical procedures may be helpful in other contexts but in this question it is
introduced before the diagnosis has been established, shifting the reasoning
trajectory from diagnostic confirmation toward treatment selection. In this
example, \textsc{PUlSE} triggers retrieval at a step more directly tied to the
missing diagnostic fact, and therefore resolves the key factual gap with fewer
retrieval calls.
\section{Ethical Considerations and Responsible Use}
\label{appendix:ethics}

\subsection{Risks and Safety}
\label{appendix:risks}

\textsc{PUlSE} is a research framework for studying factual error detection and adaptive retrieval in benchmark medical QA. Its outputs should not be interpreted as medical advice or substitutes for professional judgment. 

Potential risks include both false negatives and false positives. A false negative may allow an incorrect medical claim to influence downstream reasoning. A false positive may trigger unnecessary retrieval, introduce distracting evidence, or shift the model toward an incorrect conclusion. Retrieved passages may also be incomplete, outdated, irrelevant, or inconsistent with current medical guidance, particularly in open-world biomedical corpora. 

A low error score or retrieved passage does not guarantee factual correctness, patient-specific applicability, treatment safety, or guideline compliance. Applications beyond benchmark research would require task-specific validation, up-to-date evidence sources, and qualified professional oversight. 

\subsection{Data Privacy}
\label{appendix:privacy}

Our experiments use publicly available medical QA benchmarks and biomedical text corpora. We do not collect new patient data, access electronic health records, or process identifiable protected health information. The study is conducted in a benchmark evaluation setting and does not involve interaction with patients or human subjects.

\subsection{Use of AI Assistants}
\label{appendix:ai_use}

We use large language models to support step-level factuality annotation and data quality filtering. The prompts, label schema, and filtering procedures are described in Appendix~\ref{appendix:annotation_protocol} and Appendix~\ref{appendix:construction}. Automatically produced labels may contain errors; they serve as noisy supervision and evaluation aids for research rather than authoritative medical judgments. 
\end{document}